\documentclass{article}
\usepackage{caption}
\usepackage{subcaption} 
\usepackage{amsmath} 
\usepackage{graphicx}          
\graphicspath{{fig/}} 

\usepackage[preprint]{neurips_2025}

\usepackage[utf8]{inputenc} 
\usepackage[T1]{fontenc}    
\usepackage{hyperref}       
\usepackage{url}            
\usepackage{booktabs}       
\usepackage{amsfonts}       
\usepackage{nicefrac}       
\usepackage{microtype}      
\usepackage{xcolor}         
\usepackage{amssymb}
\usepackage{booktabs}
\usepackage{textcomp}
\usepackage{pifont}
\usepackage{geometry}
\usepackage{graphicx}

\title{\includegraphics[height=1.8ex]{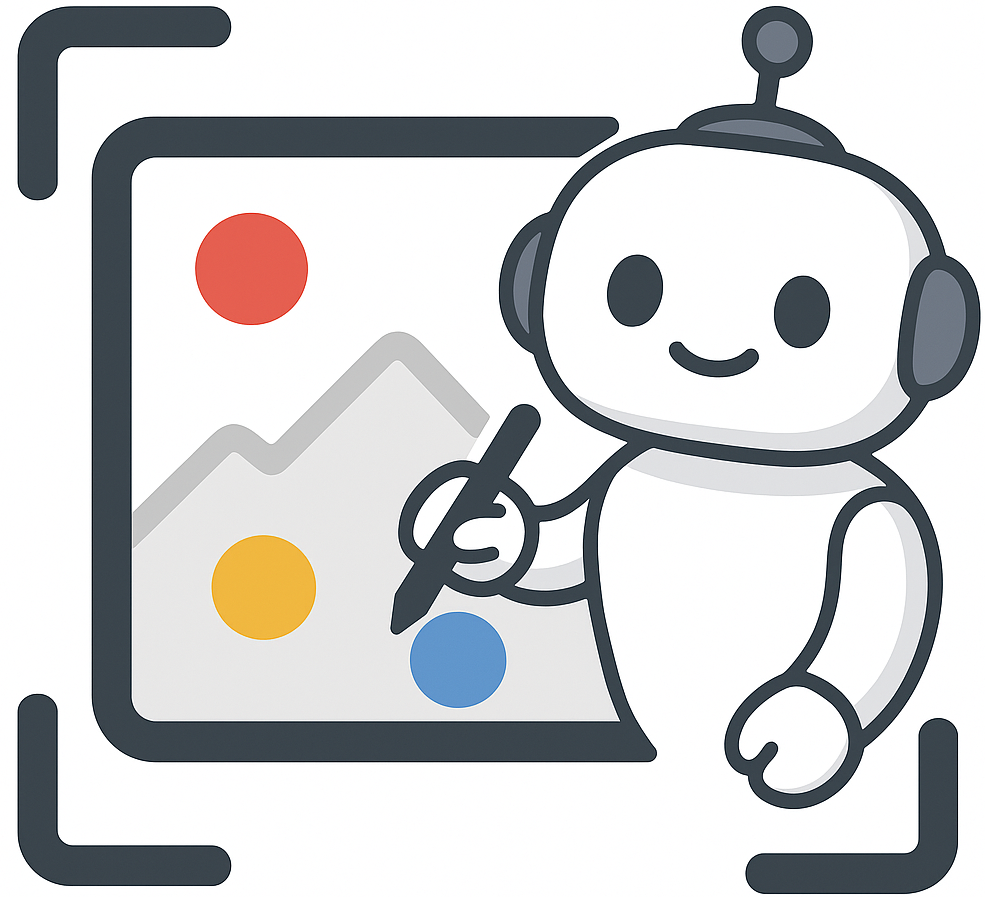} PointArena: Probing Multimodal Grounding Through Language-Guided Pointing}

\author{%
  Long Cheng$^{1}$\thanks{Co-first authors.}\quad
  Jiafei Duan$^{1,2}$\footnotemark[1]\quad
  Yi Ru Wang$^{1}$\thanks{Co-second authors.}\quad
  Haoquan Fang$^{1}$\footnotemark[2]\quad
  Boyang Li$^{1}$\footnotemark[2]\\\bf
  Yushan Huang$^{1}$\quad
  Elvis Wang$^{3}$\quad
  Ainaz Eftekhar$^{1,2}$\quad
  Jason Lee$^{1}$\quad
  Wentao Yuan$^{1}$\\\bf
  Rose Hendrix$^{2}$\quad
  Noah A.~Smith$^{1,2}$\quad
  Fei Xia$^{1}$\quad
  Dieter Fox$^{1}$\quad
  Ranjay Krishna$^{1,2}$\\[3pt]
  $^{1}$University of Washington\qquad
  $^{2}$Allen Institute for Artificial Intelligence\\
  $^{3}$Anderson Collegiate Vocational Institute\\
  \url{https://pointarena.github.io}
}

\begin{document}

\maketitle

\begin{abstract}

Pointing serves as a fundamental and intuitive mechanism for grounding language within visual contexts, with applications spanning robotics, assistive technologies, and interactive AI systems. While recent multimodal models have started to support pointing capabilities, existing benchmarks typically focus only on referential object localization tasks. We introduce \textbf{PointArena}, a comprehensive platform for evaluating multimodal pointing across diverse reasoning scenarios. PointArena comprises three components: (1) \textbf{Point-Bench}, a curated dataset containing approximately 1,000 pointing tasks across five reasoning categories; (2) \textbf{Point-Battle}, an interactive, web-based arena facilitating blind, pairwise model comparisons, which has already gathered over 4,500 anonymized votes; and (3) \textbf{Point-Act}, a real-world robotic manipulation system allowing users to directly evaluate multimodal model pointing capabilities in practical settings. We conducted extensive evaluations of both state-of-the-art open-source and proprietary multimodal models. Results indicate that \texttt{Molmo-72B} consistently outperforms other models, though proprietary models increasingly demonstrate comparable performance. Additionally, we find that supervised training specifically targeting pointing tasks significantly enhances model performance. Across our multi-stage evaluation pipeline, we also observe strong correlations, underscoring the critical role of precise pointing capabilities in enabling multimodal models to effectively bridge abstract reasoning with concrete, real-world actions.

\end{abstract}

\section{Introduction}


Pointing focuses our attention.
It is one of the earliest and most universal non-verbal methods we use to communicate intent; in fact, children learn to point as a prelinguistic form of communication~\cite{tomasello2007new}. 
Precise spatial grounding—\textit{pointing}—enables a wide range of practical and high-impact applications across robotics, assistive technology, human-computer interaction, and vision-language interfaces. In robotics, a pointing-capable model can interpret language commands like ``pick up the red cup next to the bowl” and translate them into precise spatial actions~\cite{yuan2025robopoint}, enabling fine-grained object manipulation in cluttered environments~\cite{duan2024manipulate}. 
In assistive technologies, systems can help visually impaired users by answering spatial queries such as ``where is the handle on this door?''~\cite{bigham2010vizwiz} or `which one is the garlic?' In education or creative tools, pointing allows for interactive visual tutoring, such as identifying components in scientific diagrams or guiding a learner through a painting~\cite{hu2024visual}. Even in everyday virtual assistants or search engines, the ability to refer to specific image regions via pointing could make multimodal interactions more intuitive and expressive~\cite{deitke2024molmo}. Across these domains, pointing provides a low-bandwidth yet powerful spatial interface for grounding language in vision—precise enough for manipulation~\cite{ray2024sat}, intuitive enough for communication, and general enough to scale with modern multimodal models.

\begin{figure}[htbp]
    \centering
    \includegraphics[width=\linewidth]{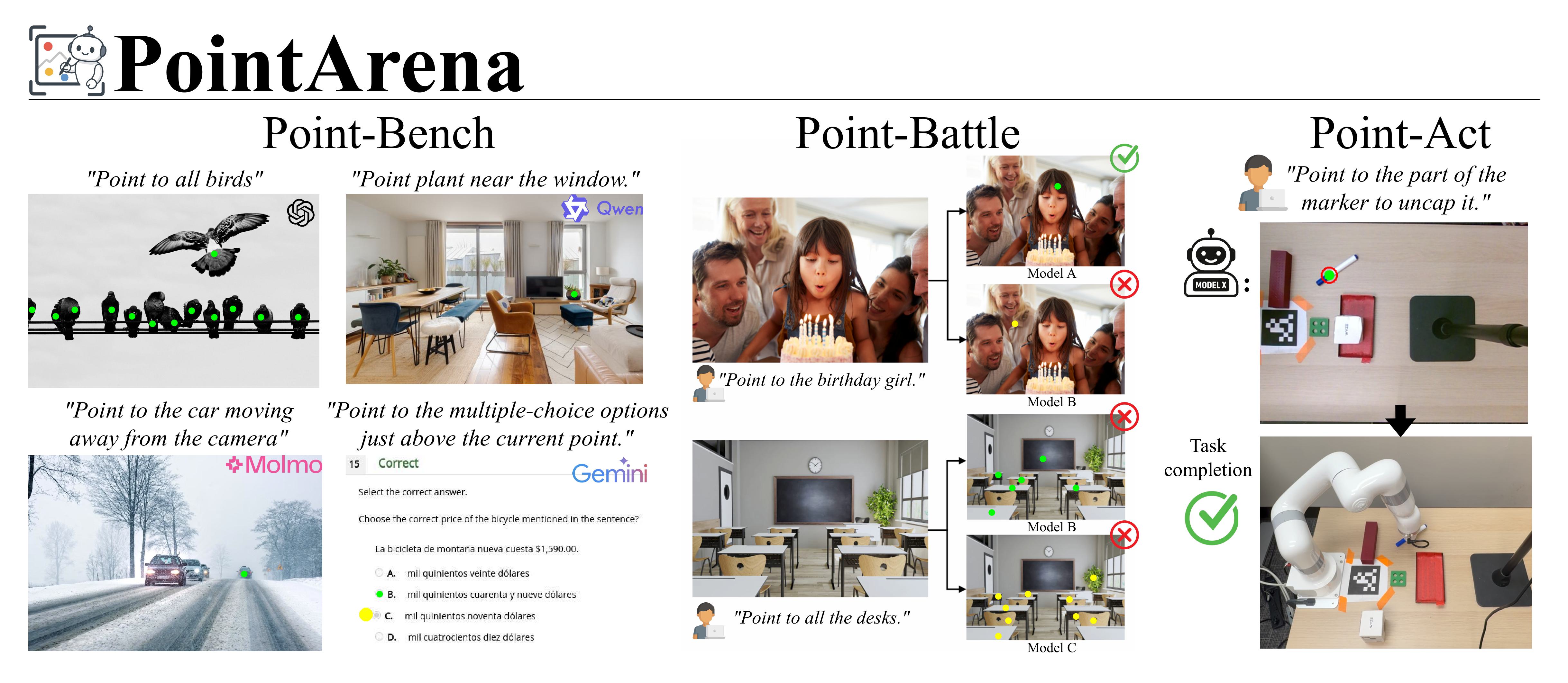}
    \caption{\textbf{Overview of \textit{PointArena}.} PointArena consists of three components: \textbf{Point-Bench}, a curated dataset for evaluating grounded pointing across five reasoning types; \textbf{Point-Battle}, a live platform for blind, pairwise model comparisons with user voting; and \textbf{Point-Act}, real-world task involving manipulation via pointing-based language commands.}
    \label{fig:f1}
\end{figure}

Recent advances in multimodal models have begun to incorporate more dynamic and spatially expressive forms of interaction. The Segment Anything Model (SAM) \cite{kirillov2023segment} enables segmentation from sparse visual prompts such as points or boxes, revealing the potential of fine-grained spatial control. Google's Gemini models \cite{georgiev2024gemini} push the boundaries of long-context visual reasoning, incorporating multiple modalities over extended sequences. In parallel, new datasets have emerged to support explicit spatial referencing. Molmo's PixMo dataset \cite{deitke2024molmo} introduces 2D pointing as a form of multimodal alignment between images and instructions, while RoboPoint \cite{yuan2025robopoint} focuses on spatial affordance prediction by linking instructions to interaction-relevant keypoints in robotic contexts. 
These setups bias evaluations toward pixel-level accuracy rather than conceptual reasoning and often lack diversity or scalability.

There is a need for a holistic evaluation platform to make progress towards language-guided pointing.
Although datasets for referring expressions exist (e.g.RefCOCO, RefCOCO+, and RefCOCOg~\cite{kazemzadeh2014referitgame,yu2016modeling}), they are focused on a subset of pointing tasks: object location.
They lack the ambiguity and contextual variability that users expect from modern interactive models, limiting their utility for studying pragmatic or interactive applications. As a result, they offer only partial insight into the full spectrum of grounding required for embodied or assistive agents.

We, therefore, propose \textbf{PointArena}, a platform to probe and evaluate grounded visual reasoning with pointing. PointArena presents a suite of tasks where a multimodal model must answer questions or resolve instructions by combining language and pointing gestures to identify specific image regions. These tasks go beyond traditional VQA by requiring spatial outputs (e.g., selecting a location or region) rather than purely textual ones. PointArena allows for both unambiguous and ambiguous scenarios, supporting studies of disambiguation, spatial commonsense, and pragmatic inference.
Unlike bounding boxes, segmentation masks, or free-form text responses, pointing offers high-precision signal that avoids reliance on object contours or dense annotations and is directly compatible with human evaluation.

\textbf{PointArena} decomposes pointing into three stages of evaluation: 1) \textbf{Point-Bench} is a curated dataset of $982$ manually selected, annotated, and verified image-question pairs across five high-level categories (Spatial, Affordance, Counting, Steerable, and Reasoning). 2) \textbf{Point-Battle} is an interactive, online platform for blind, pairwise comparison between models based on user instructions. Users select from curated or custom-uploaded images. Voting is anonymized, and we have collected over $4,500$ votes from more than $100$ participants. 3) \textbf{Point-Act} is a real-world benchmark that evaluates the utility of pointing in for a downstream application. The system directs a robotic arm to manipulate objects through pointing-based language commands. All three evaluation stages require minimal human effort; each is self-contained and can run live to evaluate any model.

Through our evaluation of both open-source and proprietary models across the three stages of the PointArena benchmark, we find that \texttt{Molmo-72B} achieves the highest performance on Point-Bench, with proprietary models such as \texttt{Gemini-2.5-Pro} performing comparably. Models trained with explicit pointing supervision consistently outperform those without. We also observe a strong correlation between static benchmark accuracy and human preference in Point-Battle. Notably, we find that adding language reasoning (e.g., Chain-of-Thought~\cite{wei2022chain}) does not improve visual grounding for pointing tasks. Our study further reveals several other actionable insights into model behavior and evaluation design. We see \textbf{PointArena} as a missing component necessary as we develop general-purpose vision-language models that can reason about and interact with the world.
\section{Related work}

\noindent\textbf{Grounding benchmarks.}
There are an abundance of benchmarks for visual grounding and spatial reasoning capabilities of multimodal large language models (MLLMs). The RefCOCO, RefCOCO+, and RefCOCOg datasets focus on 2D visual grounding, with RefCOCOg emphasizing long-form referring expressions and fine-grained object distinctions~\cite{yu2016modelingcontextreferringexpressions}. In 3D, ScanRefer provides 51,583 descriptions across 800 RGB-D scans, supporting joint language-geometry models for 3D object localization~\cite{chen2020scanrefer3dobjectlocalization}. ReferIt3D and CityRefer extend this to fine-grained and outdoor settings, with CityRefer incorporating geographic features from OpenStreetMap~\cite{achlioptas2020referit_3d, miyanishi2023cityrefergeographyaware3dvisual}. Interactive benchmarks such as GuessWhat?! evaluate multi-turn object grounding through binary dialog across 150K games~\cite{devries2017guesswhatvisualobjectdiscovery}. Flickr30K Entities supports phrase-region grounding through 276K bounding boxes and cross-caption coreference annotations~\cite{plummer2016flickr30kentitiescollectingregiontophrase}. These datasets primarily focus on bounding box grounding, object retrieval, or dialog-based localization. They do not explicitly evaluate pointing behavior. 


\noindent\textbf{Arena style evaluation.}
Arena-style evaluations have become a common framework for comparing large language models (LLMs) via pairwise comparisons and user voting. \textit{Chatbot Arena} established the core methodology, using anonymized head-to-head comparisons and Elo ratings, collecting over 240K human votes across 50+ models and 100+ languages~\cite{chiang2024chatbotarenaopenplatform}. The MT-Bench framework enhanced this approach by introducing a specialized multi-turn dialogue evaluation system that assesses contextual understanding and reasoning through AI-powered scoring~\cite{zheng2023judgingllmasajudgemtbenchchatbot}. \textit{am-ELO} addressed instability in Elo rankings by incorporating maximum likelihood estimation and annotator reliability modeling~\cite{liu2025amelostableframeworkarenabased}. \textit{Auto-Arena} fully automated the evaluation process using LLM-generated questions, peer model battles, and LLM-based voting, achieving 92.14\% agreement with human votes~\cite{zhao2024autoarenaautomatingllmevaluations}. \textit{BenchBuilder} introduced automated benchmark construction from crowdsourced data, yielding Arena-Hard-Auto with tighter confidence intervals than MT-Bench. Specialized variants such as \textit{Werewolf Arena} evaluated models on social reasoning~\cite{bailis2024werewolfarenacasestudy}, while \textit{OpenArena} enabled offline evaluations using LLMs as judges~\cite{openarena2024}. Other work has highlighted vulnerabilities, including susceptibility to vote manipulation~\cite{min2025improvingmodelrankingchatbot}, and proposed alternatives like \textit{Tournament Evaluation} to improve robustness~\cite{kelley2025tournament}. While arena-style methods are scalable and align with user preferences, they lack ground-truth supervision and are susceptible to adversarial influence. PointArena addresses these limitations by combining ground-truth evaluation (Point-Bench) with arena-style human evaluations (Point-Battle).

\noindent\textbf{Models that point or sketch.}
Multimodal large language models (MLLMs) have shown significant advances in reasoning capabilities, exemplified by GPT-4V, which demonstrates complex vision-language tasks such as image-based story generation and OCR-free mathematical reasoning~\cite{10.1093/nsr/nwae403}. Typical MLLM architectures employ encoder-decoder frameworks featuring early, intermediate, or joint visual-language fusion methods to enhance cross-modal integration. For instance, MiniGPT-4 leverages ViT-based visual features aligned with language models such as Vicuna via Q-Formers and learned projection layers~\cite{wang2024comprehensivereviewmultimodallarge}. Molmo notably improves 2D spatial grounding by predicting normalized coordinates directly from modality-specific embeddings, achieving up to 92\% precision for tasks like desktop icon localization~\cite{deitke2024molmopixmoopenweights}. \textit{RoboPoint}\cite{yuan2024robopoint} uses instruction-tuned vision-language models trained on synthetic data to predict robotic affordances, significantly outperforming GPT-4o and PIVOT\cite{nasiriany2024pivotiterativevisualprompting} by more than 20\% in spatial accuracy, and effectively supports tasks in manipulation, navigation, and augmented reality without requiring real-world supervision. Other models, such as VisCPM and Qwen-VL, incorporate region-conditioned controls and multilingual capabilities through multi-stage training approaches~\cite{10.1093/nsr/nwae403}. Additionally, newer systems like NExT-GPT extend capabilities further by integrating 3D point clouds, audio, and video modalities~\cite{10.1093/nsr/nwae403}. Despite these advancements, significant opportunities remain to further enhance the precision and accuracy of spatial localization within MLLMs. Key open questions include identifying factors that improve pointing capabilities and understanding their underlying mechanisms in multimodal models.
\section{PointArena}


Evaluating the ability of MLLMs to localize language-referred entities in images requires benchmarks that are both precise and diagnostic. Existing benchmarks often emphasize classification or captioning, but fall short when it comes to assessing fine-grained spatial grounding—the ability to resolve natural language instructions into specific image coordinates. This capability is critical not only for understanding model alignment with human intent, but also for enabling downstream applications in robotics~\cite{yuan2024robopoint}, augmented reality~\cite{duan2023ar2}, and interactive web agents~\cite{gur2023real}, and potentially contributing to explainability~\cite{park2018multimodalexplanationsjustifyingdecisions}.  As both specialized pointing models and general-purpose MLLMs improve, standardized evaluation across open-source and proprietary systems becomes essential.

We introduce \textbf{PointArena}, an evaluation suite for language-conditioned pointing. It consists of three phases of evaluation: (i) \textbf{Point-Bench}, a curated dataset for controlled measurement of spatial localization accuracy; (ii) \textbf{Point-Battle}, a live, blinded human-preference arena for pairwise model comparison; and (iii) \textbf{Point-Act}, a real-world robotic task setting that evaluates pointing precision through real-world execution. Together, these components provide a unified framework to quantify and analyze how well MLLMs ground language into visual space and even further, into the physical world.

\subsection{Task formulation}

We formalize pointing as a language-conditioned fine-grained localization task. The input consists of an RGB image \(I \in \mathbb{R}^{H \times W \times 3}\) and a natural-language instruction prompt \(q = \{w_t\}_{t=1}^{T}\).  
A multimodal large language model (MLLM) \(\mathcal{F}_{\theta}\) takes \((I, q)\) as input and predicts a set of image-space coordinate points \(P = \{(x_i, y_i)\}_{i=1}^{K}\), where each point lies within the image bounds: \(x_i \in [0, W{-}1]\), \(y_i \in [0, H{-}1]\).

Ground-truth supervision is provided as a set of binary masks \(\{M_j\}_{j=1}^{K^*}\), with each mask \(M_j \in \{0,1\}^{H \times W}\) denoting the valid region for one of \(K^*\) annotated targets. A predicted point \((x_i, y_i)\) is considered correct if it lies within the spatial support of some mask \(M_j\), i.e., \(M_j[y_i, x_i] = 1\).

A prediction is considered \emph{successful} if:
\begin{enumerate}
    \item The number of predicted points matches the number of target regions: \(K = K^*\),
    \item Each target region \(M_j\) is covered by at least one predicted point: \(\exists (x_i, y_i) \in P \;\text{such that}\; M_j[y_i, x_i] = 1\).
\end{enumerate}

This formulation enables fully automated evaluation given access to the ground-truth masks, with no need for human annotators at test time.

\begin{figure}[htbp]
    \centering
    \includegraphics[width=\linewidth]{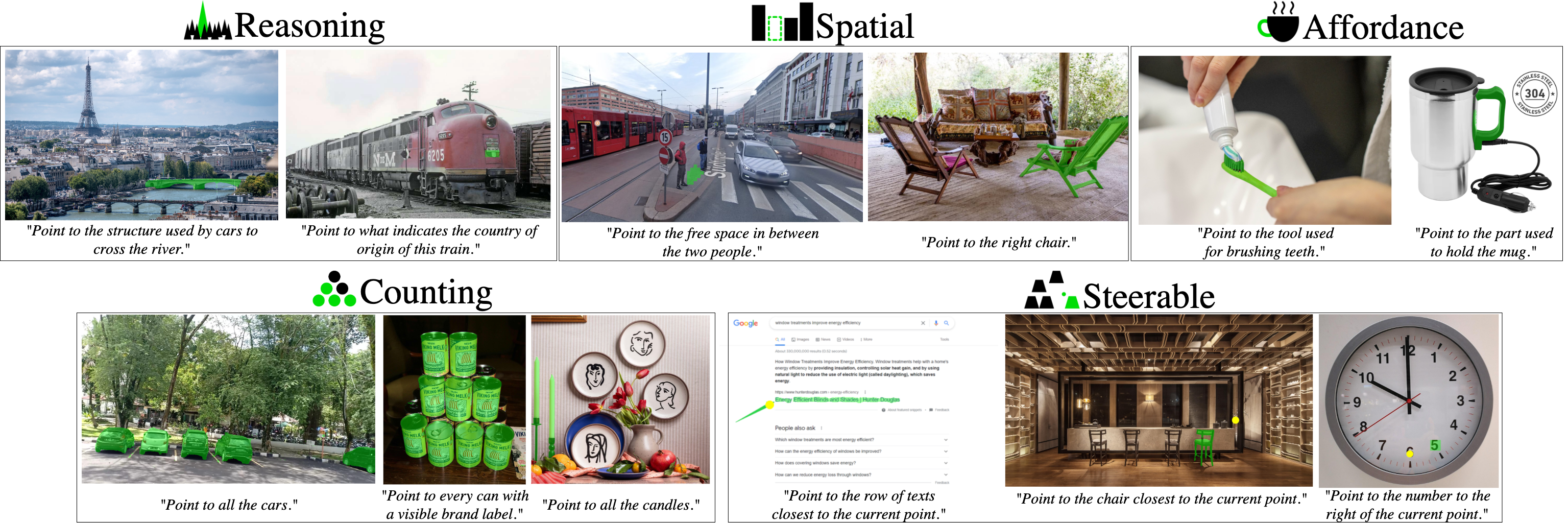}
    \caption{\textbf{Overview of the five Point-Bench categories and the annotation UI.} 
Point-Bench includes 982 image-query pairs grouped into five categories: \textbf{Spatial} (positional references), \textbf{Affordance} (functional part identification), \textbf{Counting} (attribute-based grouping), \textbf{Steerable} (relative pointing), and \textbf{Reasoning} (open-ended visual inference). Each example shows a representative query and the corresponding target. On the right, we show the Gradio-based annotation interface used to collect and refine segmentation masks. Initial masks are generated using SAM and refined by annotators, followed by manual verification.}

    \label{fig:f2}
\end{figure}

\subsection{Point-Bench}
Point-Bench is the largest benchmark for evaluating language-guided pointing, offering 982 text-image pairs with pixel-level target masks collected from public sources posted after 20 April 20, 2025. The data set is evenly divided into five task-driven categories, Spatial, Affordance, Counting, Steerable, and Reasoning, identified through a survey of the question types most frequently tackled by open-source MLLMs ~\cite{deitke2024molmopixmoopenweights,yuan2024robopoint,geminiroboticsteam2025geminiroboticsbringingai}. Annotators recruited through crowd-sourcing were free to ask any question, but category-specific image selection naturally steers them toward prompts along the desired axis (e.g., scenes with repeated objects for counting). Together, these curated splits enable systematic evaluation of an MLLM’s ability to recognize, reason, and precisely ground language in visual space.

\noindent\textbf{Spatial} — Scenes are selected for rich spatial relationships or repeated objects. Annotators craft purely positional queries; e.g., ``Point to the leftmost tree in the image.''

\noindent\textbf{Affordance} — Scenes show tabletop objects or close-ups emphasizing functional parts. Annotators ask about actionable components; e.g., ``Point to the handle used for pouring''.

\noindent\textbf{Counting} — Scenes feature multiple similar objects in varying quantities. Annotators pose queries that select a subset by number, or attribute; e.g., ``Point to all the blue cars in the image''.

\noindent\textbf{Steerable} — Images from the PixMo dataset ~\cite{deitke2024molmopixmoopenweights} each include a reference point. Annotators pose queries \emph{relative} to that point, avoiding explicit object names; e.g., ``Point to the item closest to the marked point.''

\noindent\textbf{Reasoning} — Generic, event-rich scenes invite open-ended queries that require visual reasoning, with answers conveyed by pointing; e.g., ``Point to the tallest man-made object in the image''.

To construct a Point-Bench, we developed an intuitive Gradio-based annotation interface. Annotators were shown images sampled from each category and asked to write natural language queries aligned with the category theme. These queries were then evaluated using predictions from three anonymized MLLMs. If one or fewer models produced a correct prediction as judged by human evaluators, the query was considered sufficiently challenging and accepted for inclusion in the dataset. Following this, the annotators used the same interface to annotate the target points directly on the image. A SAM model was used to generate initial masks based on the selected point, and users could refine these masks by editing or removing portions before submission. Finally, a separate group of annotators manually verified the masks to ensure they accurately reflected the user-generated queries.

\subsection{Point-Battle}
As MLLMs increasingly incorporate visually grounded reasoning and pointing capabilities, static benchmarks become inadequate for evaluating performance in open-ended, real-world scenarios—particularly with respect to human preferences. To address this limitation, we introduce \textbf{Point-Battle}, a dynamic platform for pairwise evaluation of MLLMs' pointing abilities based on user-provided language instructions. Point-Battle adopts a head-to-head evaluation format inspired by Chatbot Arena~\cite{chiang2024chatbotarenaopenplatform}, implemented through a Gradio-based web interface. In each round, two anonymized models are randomly sampled from the top performers in \textbf{Point-Bench}—including \texttt{GPT-4o}, \texttt{Gemini 2.5 Flash}, \texttt{Molmo-7B-D}, \texttt{Qwen2.5-VL-7B}, and \texttt{Grok-2 Vision}. Users submit a natural language instruction and select an image from a curated dataset (post-April 20, 2025) or upload one of their own. The two models return point predictions, which are displayed side-by-side. Participants vote for the better output or select ``both good'' or ``both bad'' if applicable. No preset prompts are provided, encouraging diverse and unbiased instructions. Model identities were kept anonymous to prevent bias. Since its launch, Point-Battle has collected over 4,500 votes from approximately 100 participants worldwide. Unlike the static \textbf{Point-Bench}, which may be subject to overfitting if used during model development, Point-Battle serves as a continuously updated benchmark that captures real-time human preferences and tracks progress in visually grounded reasoning across MLLMs. Furthermore, as Point-Battle scales up, this would also be a platform for collecting pointing data.

\subsection{Point-Act}

The first two stages of Point Arena evaluate MLLMs’ pointing capabilities through quantitative metrics and human preference assessments. However, pointing is only meaningful insofar as it enables real-world utility. To evaluate such support, we introduce \textit{Point-Act}—an interactive system where users issue natural-language instructions via a GUI to a double-blind MLLM. The model generates one or more predicted points, which are translated into actionable commands for an \texttt{xArm 6 Lite} robot. The robot executes a pick-or-place action at the indicated location using depth sensing for spatial reasoning. This setup operationalizes pointing into end-to-end physical manipulation, bridging language grounding with robotic control. Point-Act highlights the downstream consequences of grounding precision: even small localization errors can cause execution failures, whereas accurate predictions enable consistent real-world success.

\section{Experiments}


\begin{figure}[htbp]
    \centering
    \includegraphics[width=\linewidth]{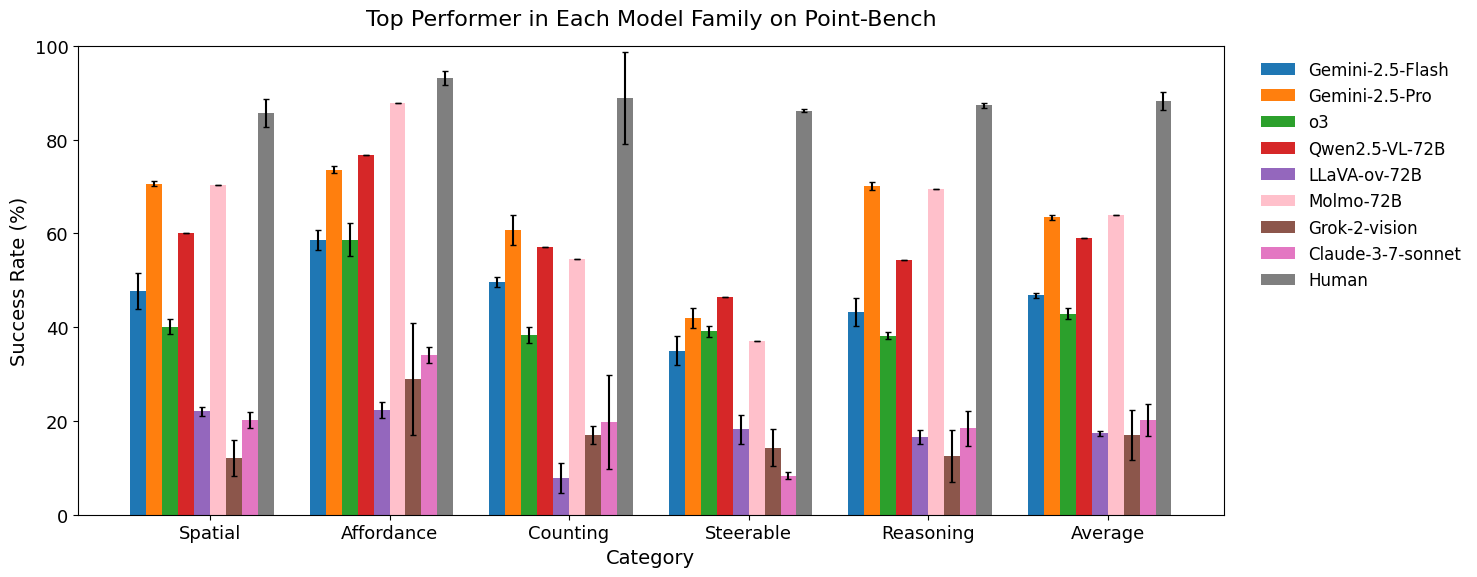}
    \caption{
        \textbf{Success rates of MLLMs on Point-Bench} across six task categories: \textit{Spatial}, \textit{Affordance}, \textit{Counting}, \textit{Steerable}, \textit{Reasoning}, and \textit{Average}. Each bar represents the mean success rate (\%) for a given model, with error bars indicating standard deviation across three evaluation runs. The ``Human'' bar serves as an upper-bound reference. The results demonstrate substantial performance disparities, with top models (e.g., GPT-4o, Gemini-2.5-Pro, Molmo-72B) achieving near-human accuracy in select categories, while others (e.g., LLaVA, Grok, and Claude) consistently underperform.
    }
    \label{fig:f3}
\end{figure}

We evaluate a range of multimodal large language models (MLLMs)—both proprietary and open-source—using three components: \textbf{Point-Bench} (static benchmark evaluation), \textbf{Point-Battle} (human preference comparison), and \textbf{Point-Act} (real-world robotic execution). Section~\ref{sec:eval_setup} describes the evaluation protocols, including model selection, prompting, and success metrics. Section~\ref{sec:main_results} presents results on model performance and the impact of pointing supervision. Section~\ref{sec:further_analysis2} presents results demonstrating the correlation between benchmark accuracy, human preference judgments, and real-world task performance in pointing tasks. Section~\ref{sec:further_analysis} includes ablations on prompt structure and output formats using GPT-4o to analyze factors affecting pointing accuracy.


\subsection{Evaluation Setup}
\label{sec:eval_setup}
All evaluations were performed under zero-shot prompting conditions. To ensure consistent outputs across models with differing internal coordinate systems—particularly proprietary ones—we adopted a standardized output format: \([x, y]\), where \(x\) and \(y\) denote horizontal and vertical pixel coordinates, respectively. This format was used across all models, except for those like \texttt{Molmo}, \texttt{Qwen2.5-VL}, and \texttt{Gemini}, which provide explicit coordinate outputs or prompting instructions.

Success was measured using a binary metric: a prediction was considered correct if the point lay within the target mask. For non-counting tasks, models were prompted to predict a single point; if multiple were returned, only the first point was evaluated, assuming it reflected the highest-confidence prediction due to the autoregressive generation process.

\vspace{0.5em}
\noindent\textbf{Point-Bench.} We benchmarked 16 MLLMs (spanning open-source and proprietary models, including key variants). Each model was evaluated on the same 982 image-instruction pairs, three times independently, to compute means and standard deviations. Open-source models were executed locally on NVIDIA A100 GPUs, while proprietary models were accessed via public APIs.

\vspace{0.5em}
\noindent\textbf{Point-Battle.} To measure alignment with human preferences, we released a live evaluation platform and promoted it via social media and mailing lists. Users voted on head-to-head comparisons between anonymous model outputs. Elo ratings were computed from pairwise comparisons excluding ambiguous votes (“both good” or “both bad”).

\vspace{0.5em}
\noindent\textbf{Point-Act.} We recruited 10 remote participants to interact with our real-world robot setup. For a fixed scene, participants evaluated three agents—\texttt{Molmo-7B-D}, \texttt{GPT-4o}, and a human reference—across three trials. After each condition, they completed a System Usability Scale (SUS) survey.

\vspace{0.5em}
\noindent\textbf{Models.} We evaluate variants from Molmo~\cite{deitke2024molmo}, Gemini~\cite{geminiroboticsteam2025geminiroboticsbringingai}, OpenAI~\cite{achiam2023gpt}, Claude~\cite{anthropic2024claude3}, Grok~\cite{xai2024grok2}, LLaVA~\cite{li2024llava}, and Qwen~\cite{bai2025qwen2}. See appendix for details.

\begin{figure}[htbp]
    \centering
    \includegraphics[width=\linewidth]{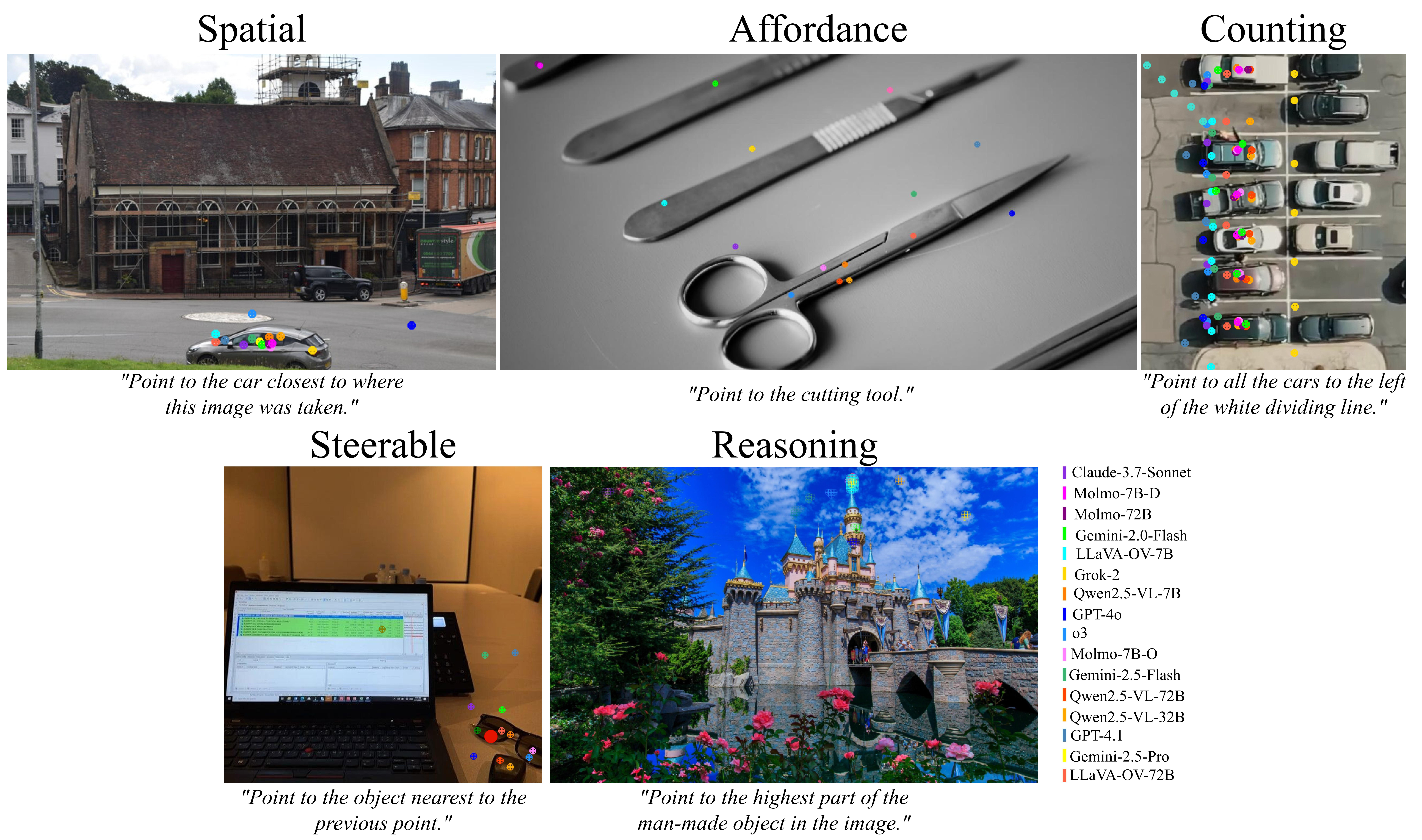}
    \caption{\textbf{Qualitative predictions across Point-Bench categories.} 
Example model predictions are shown for each of the five Point-Bench categories: \textbf{Spatial}, \textbf{Affordance}, \textbf{Counting}, \textbf{Steerable}, and \textbf{Reasoning}. Each colored dot corresponds to a prediction from a different MLLM, labeled by model name in the legend. These examples highlight the diversity of pointing behaviors and the variation in performance across models.}

    \label{fig:f4}
\end{figure}

\subsection{Main Results}
\label{sec:main_results}
\begin{figure}[htbp]
    \centering
    \includegraphics[width=\linewidth]{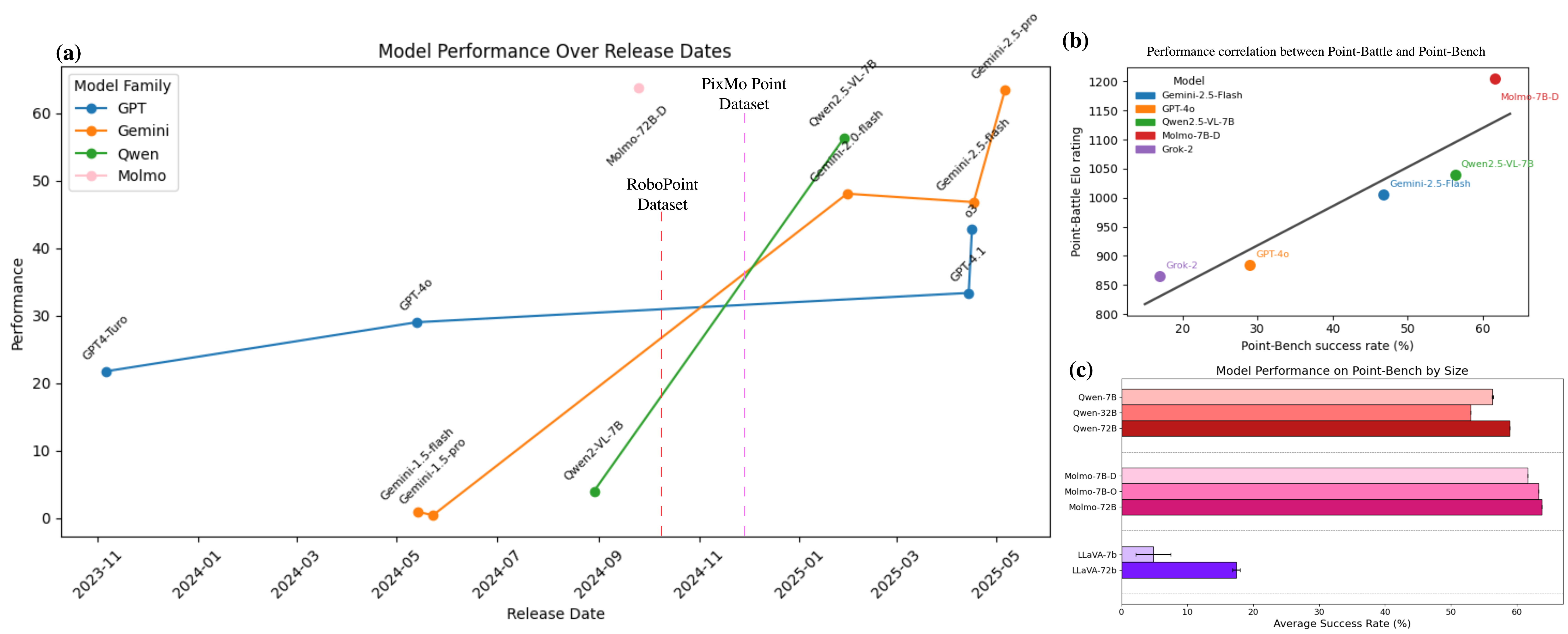}
    \caption{
\textbf{Insights drawn from Point-Battle and Point-Bench.}  
(a) This figure shows Point-Bench performance (\%) of MLLMs over time, grouped by model family. A sharp performance increase is observed in models released after the PixMo dataset or RoboPoint (dashed line, October 2024 or December 2024). Notably, \texttt{GPT4.1} improves by 21.1 percentage points over \texttt{GPT-4-Turbo}, and \texttt{Gemini-2.0-Flash} improves by 45.9 points over \texttt{Gemini-1.5-Flash}. These trends suggest that newer proprietary models may incorporate pointing supervision, potentially derived from or inspired by RoboPoint or PixMo. (b) Linear regression of the five models common to Point-Battle and Point-Bench reveals a strong correlation (\(R^{2}=0.85\)), confirming close agreement between the two evaluation frameworks. (c) Performance of open-source models as a function of parameter count. While there is a slight upward trend, the performance gains with increasing model size are marginal, suggesting diminishing returns and limited sensitivity to scale within this range.
}

    \label{fig:study1}
\end{figure}
\textbf{Open-source models perform comparably to proprietary models in pointing accuracy.} Point-Bench results show that open-source MLLMs explicitly trained on pointing data often match or outperform proprietary models. For example, \texttt{Molmo-72B} outperformed \texttt{Gemini-2.5-Pro} by 0.43 percentage points—a statistically insignificant margin ($p \approx 0.29$). In affordance reasoning, open-source models like \texttt{Molmo-72B} and \texttt{Qwen2.5-VL} consistently exceed proprietary baselines. Overall, Molmo-72B achieves the highest performance on the Point-Bench benchmark as shown in Figure \ref{fig:f3}.

\textbf{Pointing supervision significantly boosts performance.} Access to explicit pointing data is a key driver of model accuracy as shown in Figure \ref{fig:study1}a. Within the Qwen family, incorporating the PixMo corpus into \texttt{Qwen2.5-VL-7B} increased performance to 52.3\%, a substantial gain over the 17.4\% achieved by \texttt{Qwen2-VL-7B}, which did not use such data. In contrast, LLaVA variants—also trained without explicit pointing supervision—achieved only 4.8–17.4\% on average.

\textbf{Proprietary models likely benefit from open-source pointing datasets.}  
While proprietary training data is opaque, we observe large performance jumps in models released shortly after the PixMo ~\cite{deitke2024molmopixmoopenweights} and RoboPoint dataset ~\cite{yuan2024robopoint}. For instance, \texttt{GPT-o3} improved by 21.1 percentage points over \texttt{GPT-4-Turbo}, and \texttt{Gemini-2.5-Flash} improved by 45.9 points over \texttt{Gemini-1.5-Flash} (Figure~\ref{fig:study1}a). These results suggest that recent proprietary models may have incorporated PixMo or a similar corpus.

\begin{figure}[htbp]
    \centering
    \includegraphics[width=\linewidth]{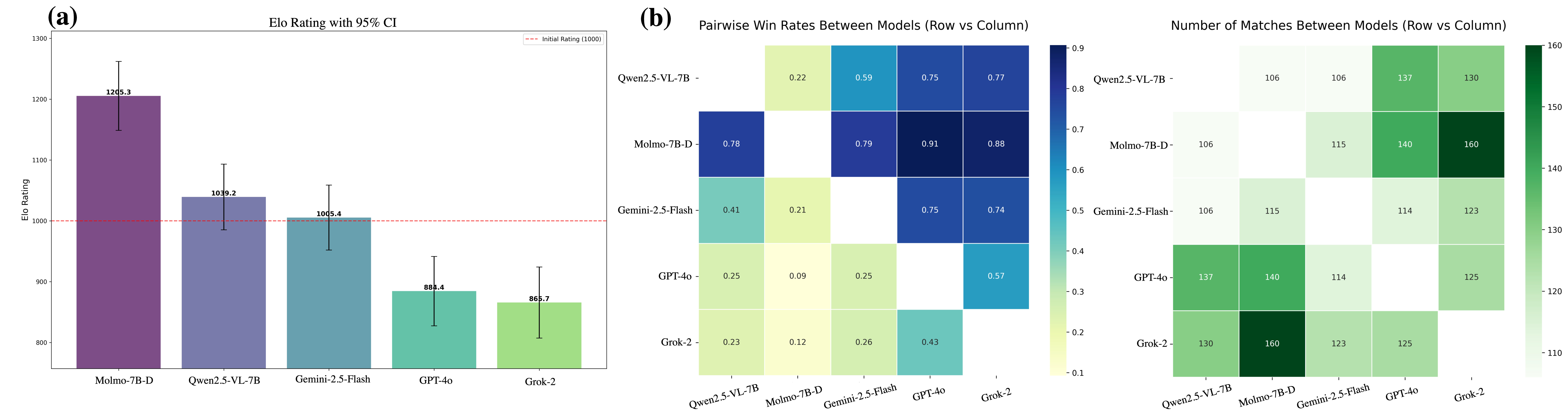}
    \caption{\textbf{Performance on Human Preference Evaluation with Point-Battle.} We collected over 4,500 votes from more than 100 global participants. Based on the Elo ratings derived from these votes, we observed a clear preference for outputs from open-source models such as \texttt{Molmo-7B-D} and \texttt{Qwen2.5-VL-7B}, which consistently outperformed proprietary models in terms of human preference.}
    \label{fig:f6}
\end{figure}

\textbf{Open-source models align more closely with human preferences.}  
In Point-Battle, \texttt{Molmo-7B-D} outperformed \texttt{Gemini-2.5-Flash} by 196 Elo points. Their 95\% confidence intervals do not overlap, and \texttt{Molmo-7B-D} won 79\% of the 115 direct head-to-head comparisons, as shown in Figure~\ref{fig:f6}a. Both \texttt{Qwen2.5-VL-7B} and \texttt{Molmo-7B-D} surpass proprietary models in human preference evaluations and exceed the 1000-point baseline, indicating a statistically significant advantage over random guessing. However, in terms of preference-aligned pointing performance, \texttt{Molmo-7B-D} remains clearly superior to \texttt{Qwen2.5-VL-7B}.


\begin{figure}[htbp]
    \centering
    \includegraphics[width=\linewidth]{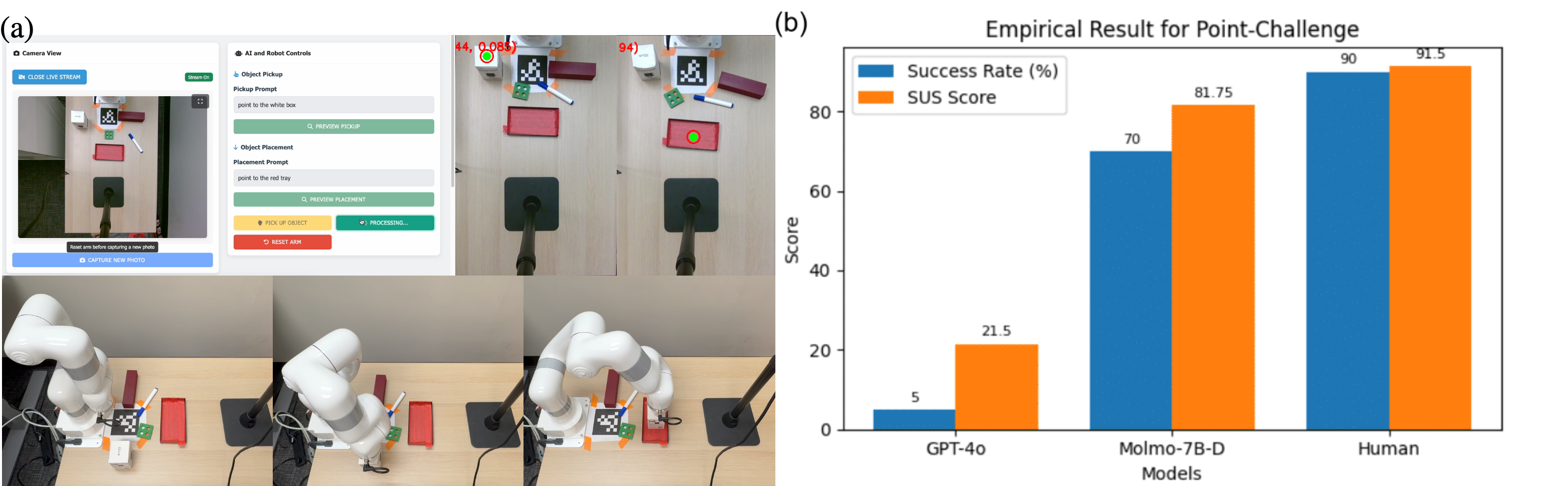}
    \caption{\textbf{Overview of the Point-Act system.} (a) The Point-Act manipulation setup enables remote control of a real-world xArm 6 Lite robot via language instructions, allowing users to evaluate pointing MLLMs. (b) User-blind evaluations and SUS preference scores collected for each model.
}
    \label{fig:f7}
\end{figure}

\textbf{Molmo excels on Point-Act evaluation.} User study results shown in Figure \ref{fig:f7} that \texttt{Molmo-7B-D} outperforms the proprietary \texttt{GPT-4o} model by a substantial margin, achieving 65\% higher performance and approaching human (oracle) baseline levels. This superiority is also reflected in user preference, with \texttt{Molmo-7B-D} scoring 60.3 points higher in SUS than \texttt{GPT-4o}.

\textbf{Model size does not impact pointing performance.} As shown in Figure~\ref{fig:study1}c, the performance of open-source models (LLaVA-OV, Molmo, and Qwen-VL) on Point-Bench remains largely unchanged with increased model size. For example, \texttt{Qwen2.5-VL-7B} performs within 3\% of \texttt{Qwen2.5-VL-72B}, and \texttt{Molmo-7B-O} differs by less than 1\% from \texttt{Molmo-72B}. These results suggest that scaling model size does not significantly improve pointing accuracy.


\subsection{Results between three evaluation frameworks.}
\label{sec:further_analysis2}

The three-stage evaluation of MLLMs’ pointing capabilities should not be viewed as isolated components, but as complementary steps in a progressive pipeline. As MLLMs improve, they are expected to advance through these stages. Therefore, understanding the correlation and agreement between stages is crucial for assessing consistent performance gains.

\textbf{Human-preference and static dataset evaluations are highly consistent.}
Point-Bench’s static dataset will inevitably plateau as MLLMs improve by training on ever-larger, real or synthetic pointing corpora (e.g., RoboPoint \cite{yuan2024robopoint}). To stay ahead, we introduce \textbf{Point-Battle}, a live arena that updates continuously and enables open-ended model comparison in real time. Validating this setup, we re-evaluated the models tested on Point-Bench and observed strong alignment: Point-Battle scores correlate with Point-Bench results at \(R^{2}=0.85\) (Figure~\ref{fig:study1}b).

\textbf{Point-Bench accuracy predicts real-world task success.}  
We validated Point-Bench as a reliable proxy by testing three agents—\texttt{Molmo-7B-D}, \texttt{GPT-4o}, and a human reference—on Point-Act. Success rates closely aligned with Point-Bench scores, yielding a strong linear correlation ($R^2 = 0.92$). This high correlation indicates that Point-Bench is a reliable proxy for the pointing capability of multimodal LLMs in practical settings.


\begin{table}[htbp]
    \centering
    \caption{Performance (\%) of different GPT-4o and Gemini-2.5-Flash variants across evaluation categories.}
    \label{tab:evaluation_results}
    \resizebox{\linewidth}{!}{%
        \begin{tabular}{lcccccc}
            \toprule
            \textbf{Method} & \textbf{Affordance} & \textbf{Spatial} & \textbf{Reasoning} & \textbf{Steerability} & \textbf{Counting} & \textbf{Average} \\
            \midrule
            GPT-4o + In-Context (2-shots)   & 46.0 & 26.7 & 23.9 & 22.5 & 33.2 & 30.4 \\
            GPT-4o + Chain-of-Thought (CoT)~\cite{wei2022chain}    & 41.4 & 24.6 & 21.2 & 13.5 & 32.1 & 26.6 \\
            GPT-4o + Unparsed language instruction     & 37.8 & 23.7 & 19.7 & 21.9 & 31.1 & 26.9 \\
            GPT-4o (Default)                   & 42.4 & 25.6 & 23.8 & 24.5 & 31.1 & 29.5 \\
            \midrule
            Gemini-2.5-Flash + In-Context (2-shots)   & 50.0 & 40.5 & 35.8 & 32.0 & 40.3 & 39.7 \\
            Gemini-2.5-Flash + Chain-of-Thought (CoT)~\cite{wei2022chain}     &44.4 & 25.6 &	24.9 &	26.0 &	33.7 &	30.9	\\
            Gemini-2.5-Flash + Unparsed language instruction     & 55.0 &	44.6 &	43.5 &	25.0 &	47.5 &	43.1 \\
            Gemini-2.5-Flash (Default)   & 58.6 & 47.7 & 43.2 & 35.0 & 49.7 & 46.8 \\
            \bottomrule
        \end{tabular}
    }
\end{table}

\subsection{What Other Factors Drive Pointing Performance?}

\label{sec:further_analysis}

To understand the design choices that impact pointing, we conducted ablations on \texttt{GPT-4o} using variations in prompt structure and output representation, as shown in Table \ref{tab:evaluation_results}.


\textbf{Targeted prompts outperform verbose reasoning.}  
Incorporating Chain-of-Thought (CoT) reasoning reduced pointing accuracy by 2.9\% for \texttt{GPT-4o} and by a substantial 16\% for \texttt{Gemini-2.5-Flash}. Using raw, unfiltered user queries led to an additional drop of 2.6\% and 3.7\% for \texttt{GPT-4o} and \texttt{Gemini-2.5-Flash}, respectively. These results suggest that clear, targeted prompts with well-defined coordinate systems are crucial for effective pointing, while additional reasoning through language does not enhance MLLMs’ pointing capabilities.

\section{Limitations, Discussion, and Conclusions}

\noindent\textbf{Discussions.}  
PointArena integrates static benchmarks through Point-Bench and human preference-based comparisons via Point-Battle to evaluate spatial reasoning in multimodal models. To improve annotation fidelity, future pipelines may replace the current grid-based refinement tools with free-form contouring interfaces, allowing annotators to directly trace object boundaries with a mouse or stylus. This could yield smoother, more precise masks—particularly around object edges, where coarse grids often fail. To address benchmark staleness, we augment Point-Bench with user-generated content from Point-Battle, where participants upload images and provide implicit supervision through interaction. Although these annotations are noisier than manually curated ones, they enable scalable, up-to-date evaluations. Finally, we plan to implement an adaptive sampling strategy that dynamically selects model pairs with similar performance, increasing the informativeness of each comparison.

\noindent\textbf{Limitations.}  
The current annotation pipeline relies on the Segment Anything Model (SAM)~\cite{kirillov2023segment} to generate initial masks, which annotators refine through a grid-based interface. While efficient, this approach often results in coarse and imprecise boundaries, particularly for fine-grained or irregular shapes. This degrades segmentation quality and introduces noise into downstream evaluation. Additionally, as large multimodal models are frequently trained on publicly available datasets, static benchmarks such as Point-Bench are at increasing risk of becoming part of training data, reducing their effectiveness in evaluating generalization. Lastly, Point-Battle currently selects model pairs uniformly at random, which leads to uninformative comparisons—especially between models with large performance gaps—limiting the efficiency of the evaluation process.

\noindent\textbf{Conclusion.}  
PointArena offers a unified, extensible framework for benchmarking language-conditioned pointing in multimodal models, combining static evaluations with live user-driven comparisons. While current limitations include coarse annotation tools, dataset contamination risks, and inefficient evaluation strategies, our proposed improvements aim to enhance scalability, annotation fidelity, and comparative signal. As the field evolves, PointArena is designed to adapt, supporting rigorous, real-world evaluation of grounded spatial reasoning.

\begin{ack}
We gratefully acknowledge the contributions of our volunteer annotators, listed in no particular order: Rachel Edwards and Kathryn Slusarczyk. We also thank Jieyu Zhang for valuable discussions on the paper. Jiafei Duan is supported by the A*STAR National Science PhD Fellowship.

\end{ack}

\bibliographystyle{plainnat}  
\bibliography{references}

\begin{thebibliography}{40}
\providecommand{\natexlab}[1]{#1}
\providecommand{\url}[1]{\texttt{#1}}
\expandafter\ifx\csname urlstyle\endcsname\relax
  \providecommand{\doi}[1]{doi: #1}\else
  \providecommand{\doi}{doi: \begingroup \urlstyle{rm}\Url}\fi

\bibitem[Achiam et~al.(2023)Achiam, Adler, Agarwal, Ahmad, Akkaya, Aleman, Almeida, Altenschmidt, Altman, Anadkat, et~al.]{achiam2023gpt}
Josh Achiam, Steven Adler, Sandhini Agarwal, Lama Ahmad, Ilge Akkaya, Florencia~Leoni Aleman, Diogo Almeida, Janko Altenschmidt, Sam Altman, Shyamal Anadkat, et~al.
\newblock Gpt-4 technical report.
\newblock \emph{arXiv preprint arXiv:2303.08774}, 2023.

\bibitem[Achlioptas et~al.(2020)Achlioptas, Abdelreheem, Xia, Elhoseiny, and Guibas]{achlioptas2020referit_3d}
Panos Achlioptas, Ahmed Abdelreheem, Fei Xia, Mohamed Elhoseiny, and Leonidas~J. Guibas.
\newblock {ReferIt3D}: Neural listeners for fine-grained 3d object identification in real-world scenes.
\newblock In \emph{16th European Conference on Computer Vision (ECCV)}, 2020.

\bibitem[Anthropic(2024)]{anthropic2024claude3}
Anthropic.
\newblock The claude 3 model family: Opus, sonnet, haiku.
\newblock \url{https://www.anthropic.com/news/claude-3-family}, 2024.
\newblock Claude-3 Model Card.

\bibitem[Bai et~al.(2025)Bai, Chen, Liu, Wang, Ge, Song, Dang, Wang, Wang, Tang, et~al.]{bai2025qwen2}
Shuai Bai, Keqin Chen, Xuejing Liu, Jialin Wang, Wenbin Ge, Sibo Song, Kai Dang, Peng Wang, Shijie Wang, Jun Tang, et~al.
\newblock Qwen2. 5-vl technical report.
\newblock \emph{arXiv preprint arXiv:2502.13923}, 2025.

\bibitem[Bailis et~al.(2024)Bailis, Friedhoff, and Chen]{bailis2024werewolfarenacasestudy}
Suma Bailis, Jane Friedhoff, and Feiyang Chen.
\newblock Werewolf arena: A case study in llm evaluation via social deduction, 2024.
\newblock URL \url{https://arxiv.org/abs/2407.13943}.

\bibitem[Bigham et~al.(2010)Bigham, Jayant, Ji, Little, Miller, Miller, Miller, Tatarowicz, White, White, et~al.]{bigham2010vizwiz}
Jeffrey~P Bigham, Chandrika Jayant, Hanjie Ji, Greg Little, Andrew Miller, Robert~C Miller, Robin Miller, Aubrey Tatarowicz, Brandyn White, Samual White, et~al.
\newblock Vizwiz: nearly real-time answers to visual questions.
\newblock In \emph{Proceedings of the 23nd annual ACM symposium on User interface software and technology}, pages 333--342, 2010.

\bibitem[Chen et~al.(2020)Chen, Chang, and Nießner]{chen2020scanrefer3dobjectlocalization}
Dave~Zhenyu Chen, Angel~X. Chang, and Matthias Nießner.
\newblock Scanrefer: 3d object localization in rgb-d scans using natural language, 2020.
\newblock URL \url{https://arxiv.org/abs/1912.08830}.

\bibitem[Chiang et~al.(2024)Chiang, Zheng, Sheng, Angelopoulos, Li, Li, Zhang, Zhu, Jordan, Gonzalez, and Stoica]{chiang2024chatbotarenaopenplatform}
Wei-Lin Chiang, Lianmin Zheng, Ying Sheng, Anastasios~Nikolas Angelopoulos, Tianle Li, Dacheng Li, Hao Zhang, Banghua Zhu, Michael Jordan, Joseph~E. Gonzalez, and Ion Stoica.
\newblock Chatbot arena: An open platform for evaluating llms by human preference, 2024.
\newblock URL \url{https://arxiv.org/abs/2403.04132}.

\bibitem[de~Vries et~al.(2017)de~Vries, Strub, Chandar, Pietquin, Larochelle, and Courville]{devries2017guesswhatvisualobjectdiscovery}
Harm de~Vries, Florian Strub, Sarath Chandar, Olivier Pietquin, Hugo Larochelle, and Aaron Courville.
\newblock Guesswhat?! visual object discovery through multi-modal dialogue, 2017.
\newblock URL \url{https://arxiv.org/abs/1611.08481}.

\bibitem[Deitke et~al.(2024{\natexlab{a}})Deitke, Clark, Lee, Tripathi, Yang, Park, Salehi, Muennighoff, Lo, Soldaini, Lu, Anderson, Bransom, Ehsani, Ngo, Chen, Patel, Yatskar, Callison-Burch, Head, Hendrix, Bastani, VanderBilt, Lambert, Chou, Chheda, Sparks, Skjonsberg, Schmitz, Sarnat, Bischoff, Walsh, Newell, Wolters, Gupta, Zeng, Borchardt, Groeneveld, Dumas, Nam, Lebrecht, Wittlif, Schoenick, Michel, Krishna, Weihs, Smith, Hajishirzi, Girshick, Farhadi, and Kembhavi]{deitke2024molmo}
Matt Deitke, Christopher Clark, Sangho Lee, Rohun Tripathi, Yue Yang, Jae~Sung Park, Mohammadreza Salehi, Niklas Muennighoff, Kyle Lo, Luca Soldaini, Jiasen Lu, Taira Anderson, Erin Bransom, Kiana Ehsani, Huong Ngo, YenSung Chen, Ajay Patel, Mark Yatskar, Chris Callison-Burch, Andrew Head, Rose Hendrix, Favyen Bastani, Eli VanderBilt, Nathan Lambert, Yvonne Chou, Arnavi Chheda, Jenna Sparks, Sam Skjonsberg, Michael Schmitz, Aaron Sarnat, Byron Bischoff, Pete Walsh, Chris Newell, Piper Wolters, Tanmay Gupta, Kuo-Hao Zeng, Jon Borchardt, Dirk Groeneveld, Jen Dumas, Crystal Nam, Sophie Lebrecht, Caitlin Wittlif, Carissa Schoenick, Oscar Michel, Ranjay Krishna, Luca Weihs, Noah~A. Smith, Hannaneh Hajishirzi, Ross Girshick, Ali Farhadi, and Aniruddha Kembhavi.
\newblock Molmo and pixmo: Open weights and open data for state-of-the-art multimodal models.
\newblock \emph{arXiv preprint arXiv:2409.17146}, 2024{\natexlab{a}}.

\bibitem[Deitke et~al.(2024{\natexlab{b}})Deitke, Clark, Lee, Tripathi, Yang, Park, Salehi, Muennighoff, Lo, Soldaini, Lu, Anderson, Bransom, Ehsani, Ngo, Chen, Patel, Yatskar, Callison-Burch, Head, Hendrix, Bastani, VanderBilt, Lambert, Chou, Chheda, Sparks, Skjonsberg, Schmitz, Sarnat, Bischoff, Walsh, Newell, Wolters, Gupta, Zeng, Borchardt, Groeneveld, Nam, Lebrecht, Wittlif, Schoenick, Michel, Krishna, Weihs, Smith, Hajishirzi, Girshick, Farhadi, and Kembhavi]{deitke2024molmopixmoopenweights}
Matt Deitke, Christopher Clark, Sangho Lee, Rohun Tripathi, Yue Yang, Jae~Sung Park, Mohammadreza Salehi, Niklas Muennighoff, Kyle Lo, Luca Soldaini, Jiasen Lu, Taira Anderson, Erin Bransom, Kiana Ehsani, Huong Ngo, YenSung Chen, Ajay Patel, Mark Yatskar, Chris Callison-Burch, Andrew Head, Rose Hendrix, Favyen Bastani, Eli VanderBilt, Nathan Lambert, Yvonne Chou, Arnavi Chheda, Jenna Sparks, Sam Skjonsberg, Michael Schmitz, Aaron Sarnat, Byron Bischoff, Pete Walsh, Chris Newell, Piper Wolters, Tanmay Gupta, Kuo-Hao Zeng, Jon Borchardt, Dirk Groeneveld, Crystal Nam, Sophie Lebrecht, Caitlin Wittlif, Carissa Schoenick, Oscar Michel, Ranjay Krishna, Luca Weihs, Noah~A. Smith, Hannaneh Hajishirzi, Ross Girshick, Ali Farhadi, and Aniruddha Kembhavi.
\newblock Molmo and pixmo: Open weights and open data for state-of-the-art vision-language models, 2024{\natexlab{b}}.
\newblock URL \url{https://arxiv.org/abs/2409.17146}.

\bibitem[Duan et~al.(2023)Duan, Wang, Shridhar, Fox, and Krishna]{duan2023ar2}
Jiafei Duan, Yi~Ru Wang, Mohit Shridhar, Dieter Fox, and Ranjay Krishna.
\newblock Ar2-d2: Training a robot without a robot.
\newblock \emph{arXiv preprint arXiv:2306.13818}, 2023.

\bibitem[Duan et~al.(2024)Duan, Yuan, Pumacay, Wang, Ehsani, Fox, and Krishna]{duan2024manipulate}
Jiafei Duan, Wentao Yuan, Wilbert Pumacay, Yi~Ru Wang, Kiana Ehsani, Dieter Fox, and Ranjay Krishna.
\newblock Manipulate-anything: Automating real-world robots using vision-language models.
\newblock \emph{arXiv preprint arXiv:2406.18915}, 2024.

\bibitem[Georgiev et~al.(2024)]{georgiev2024gemini}
Petko Georgiev et~al.
\newblock Gemini 1.5: Unlocking multimodal understanding across millions of tokens of context.
\newblock \emph{arXiv preprint arXiv:2403.05530}, 2024.

\bibitem[Gur et~al.(2023)Gur, Furuta, Huang, Safdari, Matsuo, Eck, and Faust]{gur2023real}
Izzeddin Gur, Hiroki Furuta, Austin Huang, Mustafa Safdari, Yutaka Matsuo, Douglas Eck, and Aleksandra Faust.
\newblock A real-world webagent with planning, long context understanding, and program synthesis.
\newblock \emph{arXiv preprint arXiv:2307.12856}, 2023.

\bibitem[Hu et~al.(2024)Hu, Shi, Fu, Roth, Ostendorf, Zettlemoyer, Smith, and Krishna]{hu2024visual}
Yushi Hu, Weijia Shi, Xingyu Fu, Dan Roth, Mari Ostendorf, Luke Zettlemoyer, Noah~A Smith, and Ranjay Krishna.
\newblock Visual sketchpad: Sketching as a visual chain of thought for multimodal language models.
\newblock \emph{arXiv preprint arXiv:2406.09403}, 2024.

\bibitem[Kazemzadeh et~al.(2014)Kazemzadeh, Ordonez, Matten, and Berg]{kazemzadeh2014referitgame}
Sahar Kazemzadeh, Vicente Ordonez, Mark Matten, and Tamara Berg.
\newblock Referitgame: Referring to objects in photographs of natural scenes.
\newblock In \emph{Proceedings of the 2014 conference on empirical methods in natural language processing (EMNLP)}, pages 787--798, 2014.

\bibitem[Kelley and Wilson(2025)]{kelley2025tournament}
Richard Kelley and Duncan Wilson.
\newblock Tournament evaluation of large language models, 2025.
\newblock URL \url{https://openreview.net/forum?id=5ZpN6W5uRm}.

\bibitem[Kirillov et~al.(2023)Kirillov, Mintun, Ravi, Mao, Rolland, Gustafson, Xiao, Whitehead, Berg, Lo, Dollár, and Girshick]{kirillov2023segment}
Alexander Kirillov, Eric Mintun, Nikhila Ravi, Hanzi Mao, Chloe Rolland, Laura Gustafson, Tete Xiao, Spencer Whitehead, Alexander~C. Berg, Wan-Yen Lo, Piotr Dollár, and Ross Girshick.
\newblock Segment anything, 2023.
\newblock URL \url{https://arxiv.org/abs/2304.02643}.

\bibitem[Li et~al.(2024)Li, Zhang, Guo, Zhang, Li, Zhang, Zhang, Zhang, Li, Liu, et~al.]{li2024llava}
Bo~Li, Yuanhan Zhang, Dong Guo, Renrui Zhang, Feng Li, Hao Zhang, Kaichen Zhang, Peiyuan Zhang, Yanwei Li, Ziwei Liu, et~al.
\newblock Llava-onevision: Easy visual task transfer.
\newblock \emph{arXiv preprint arXiv:2408.03326}, 2024.

\bibitem[Liu et~al.(2025)Liu, Li, Zhuang, Liu, Shen, Ouyang, Cheng, and Wang]{liu2025amelostableframeworkarenabased}
Zirui Liu, Jiatong Li, Yan Zhuang, Qi~Liu, Shuanghong Shen, Jie Ouyang, Mingyue Cheng, and Shijin Wang.
\newblock am-elo: A stable framework for arena-based llm evaluation, 2025.
\newblock URL \url{https://arxiv.org/abs/2505.03475}.

\bibitem[Min et~al.(2025)Min, Pang, Du, Liu, Cheng, and Lin]{min2025improvingmodelrankingchatbot}
Rui Min, Tianyu Pang, Chao Du, Qian Liu, Minhao Cheng, and Min Lin.
\newblock Improving your model ranking on chatbot arena by vote rigging, 2025.
\newblock URL \url{https://arxiv.org/abs/2501.17858}.

\bibitem[Miyanishi et~al.(2023)Miyanishi, Kitamori, Kurita, Lee, Kawanabe, and Inoue]{miyanishi2023cityrefergeographyaware3dvisual}
Taiki Miyanishi, Fumiya Kitamori, Shuhei Kurita, Jungdae Lee, Motoaki Kawanabe, and Nakamasa Inoue.
\newblock Cityrefer: Geography-aware 3d visual grounding dataset on city-scale point cloud data, 2023.
\newblock URL \url{https://arxiv.org/abs/2310.18773}.

\bibitem[Nasiriany et~al.(2024)Nasiriany, Xia, Yu, Xiao, Liang, Dasgupta, Xie, Driess, Wahid, Xu, Vuong, Zhang, Lee, Lee, Xu, Kirmani, Zhu, Zeng, Hausman, Heess, Finn, Levine, and Ichter]{nasiriany2024pivotiterativevisualprompting}
Soroush Nasiriany, Fei Xia, Wenhao Yu, Ted Xiao, Jacky Liang, Ishita Dasgupta, Annie Xie, Danny Driess, Ayzaan Wahid, Zhuo Xu, Quan Vuong, Tingnan Zhang, Tsang-Wei~Edward Lee, Kuang-Huei Lee, Peng Xu, Sean Kirmani, Yuke Zhu, Andy Zeng, Karol Hausman, Nicolas Heess, Chelsea Finn, Sergey Levine, and Brian Ichter.
\newblock Pivot: Iterative visual prompting elicits actionable knowledge for vlms, 2024.
\newblock URL \url{https://arxiv.org/abs/2402.07872}.

\bibitem[Park et~al.(2018)Park, Hendricks, Akata, Rohrbach, Schiele, Darrell, and Rohrbach]{park2018multimodalexplanationsjustifyingdecisions}
Dong~Huk Park, Lisa~Anne Hendricks, Zeynep Akata, Anna Rohrbach, Bernt Schiele, Trevor Darrell, and Marcus Rohrbach.
\newblock Multimodal explanations: Justifying decisions and pointing to the evidence, 2018.
\newblock URL \url{https://arxiv.org/abs/1802.08129}.

\bibitem[Plummer et~al.(2016)Plummer, Wang, Cervantes, Caicedo, Hockenmaier, and Lazebnik]{plummer2016flickr30kentitiescollectingregiontophrase}
Bryan~A. Plummer, Liwei Wang, Chris~M. Cervantes, Juan~C. Caicedo, Julia Hockenmaier, and Svetlana Lazebnik.
\newblock Flickr30k entities: Collecting region-to-phrase correspondences for richer image-to-sentence models, 2016.
\newblock URL \url{https://arxiv.org/abs/1505.04870}.

\bibitem[Ray et~al.(2024)Ray, Duan, Tan, Bashkirova, Hendrix, Ehsani, Kembhavi, Plummer, Krishna, Zeng, et~al.]{ray2024sat}
Arijit Ray, Jiafei Duan, Reuben Tan, Dina Bashkirova, Rose Hendrix, Kiana Ehsani, Aniruddha Kembhavi, Bryan~A Plummer, Ranjay Krishna, Kuo-Hao Zeng, et~al.
\newblock Sat: Spatial aptitude training for multimodal language models.
\newblock \emph{arXiv preprint arXiv:2412.07755}, 2024.

\bibitem[SYV-AI(2024)]{openarena2024}
SYV-AI.
\newblock Openarena: An open platform for llm-as-a-judge evaluation.
\newblock \url{https://github.com/syv-ai/OpenArena}, 2024.
\newblock Accessed: 2025-05-09.

\bibitem[Team et~al.(2025)Team, Abeyruwan, Ainslie, Alayrac, Arenas, Armstrong, Balakrishna, Baruch, Bauza, Blokzijl, Bohez, Bousmalis, Brohan, Buschmann, Byravan, Cabi, Caluwaerts, Casarini, Chang, Chen, Chen, Chiang, Choromanski, D'Ambrosio, Dasari, Davchev, Devin, Palo, Ding, Dostmohamed, Driess, Du, Dwibedi, Elabd, Fantacci, Fong, Frey, Fu, Giustina, Gopalakrishnan, Graesser, Hasenclever, Heess, Hernaez, Herzog, Hofer, Humplik, Iscen, Jacob, Jain, Julian, Kalashnikov, Karagozler, Karp, Kew, Kirkland, Kirmani, Kuang, Lampe, Laurens, Leal, Lee, Lee, Liang, Lin, Maddineni, Majumdar, Michaely, Moreno, Neunert, Nori, Parada, Parisotto, Pastor, Pooley, Rao, Reymann, Sadigh, Saliceti, Sanketi, Sermanet, Shah, Sharma, Shea, Shu, Sindhwani, Singh, Soricut, Springenberg, Sterneck, Surdulescu, Tan, Tompson, Vanhoucke, Varley, Vesom, Vezzani, Vinyals, Wahid, Welker, Wohlhart, Xia, Xiao, Xie, Xie, Xu, Xu, Xu, Xu, Yang, Yao, Yaroshenko, Yu, Yuan, Zhang, Zhang, Zhou, and
  Zhou]{geminiroboticsteam2025geminiroboticsbringingai}
Gemini~Robotics Team, Saminda Abeyruwan, Joshua Ainslie, Jean-Baptiste Alayrac, Montserrat~Gonzalez Arenas, Travis Armstrong, Ashwin Balakrishna, Robert Baruch, Maria Bauza, Michiel Blokzijl, Steven Bohez, Konstantinos Bousmalis, Anthony Brohan, Thomas Buschmann, Arunkumar Byravan, Serkan Cabi, Ken Caluwaerts, Federico Casarini, Oscar Chang, Jose~Enrique Chen, Xi~Chen, Hao-Tien~Lewis Chiang, Krzysztof Choromanski, David D'Ambrosio, Sudeep Dasari, Todor Davchev, Coline Devin, Norman~Di Palo, Tianli Ding, Adil Dostmohamed, Danny Driess, Yilun Du, Debidatta Dwibedi, Michael Elabd, Claudio Fantacci, Cody Fong, Erik Frey, Chuyuan Fu, Marissa Giustina, Keerthana Gopalakrishnan, Laura Graesser, Leonard Hasenclever, Nicolas Heess, Brandon Hernaez, Alexander Herzog, R.~Alex Hofer, Jan Humplik, Atil Iscen, Mithun~George Jacob, Deepali Jain, Ryan Julian, Dmitry Kalashnikov, M.~Emre Karagozler, Stefani Karp, Chase Kew, Jerad Kirkland, Sean Kirmani, Yuheng Kuang, Thomas Lampe, Antoine Laurens, Isabel Leal, Alex~X. Lee,
  Tsang-Wei~Edward Lee, Jacky Liang, Yixin Lin, Sharath Maddineni, Anirudha Majumdar, Assaf~Hurwitz Michaely, Robert Moreno, Michael Neunert, Francesco Nori, Carolina Parada, Emilio Parisotto, Peter Pastor, Acorn Pooley, Kanishka Rao, Krista Reymann, Dorsa Sadigh, Stefano Saliceti, Pannag Sanketi, Pierre Sermanet, Dhruv Shah, Mohit Sharma, Kathryn Shea, Charles Shu, Vikas Sindhwani, Sumeet Singh, Radu Soricut, Jost~Tobias Springenberg, Rachel Sterneck, Razvan Surdulescu, Jie Tan, Jonathan Tompson, Vincent Vanhoucke, Jake Varley, Grace Vesom, Giulia Vezzani, Oriol Vinyals, Ayzaan Wahid, Stefan Welker, Paul Wohlhart, Fei Xia, Ted Xiao, Annie Xie, Jinyu Xie, Peng Xu, Sichun Xu, Ying Xu, Zhuo Xu, Yuxiang Yang, Rui Yao, Sergey Yaroshenko, Wenhao Yu, Wentao Yuan, Jingwei Zhang, Tingnan Zhang, Allan Zhou, and Yuxiang Zhou.
\newblock Gemini robotics: Bringing ai into the physical world, 2025.
\newblock URL \url{https://arxiv.org/abs/2503.20020}.

\bibitem[Tomasello et~al.(2007)Tomasello, Carpenter, and Liszkowski]{tomasello2007new}
Michael Tomasello, Malinda Carpenter, and Ulf Liszkowski.
\newblock A new look at infant pointing.
\newblock \emph{Child development}, 78\penalty0 (3):\penalty0 705--722, 2007.

\bibitem[Wang et~al.(2024)Wang, Jiang, Liu, Ma, Zhang, Pan, Liu, Gu, Xia, Li, Zhang, Wu, Liu, Zhong, Ge, Zhang, Qiang, Hu, Jiang, Zhang, Zhang, Shen, Liu, and Zhang]{wang2024comprehensivereviewmultimodallarge}
Jiaqi Wang, Hanqi Jiang, Yiheng Liu, Chong Ma, Xu~Zhang, Yi~Pan, Mengyuan Liu, Peiran Gu, Sichen Xia, Wenjun Li, Yutong Zhang, Zihao Wu, Zhengliang Liu, Tianyang Zhong, Bao Ge, Tuo Zhang, Ning Qiang, Xintao Hu, Xi~Jiang, Xin Zhang, Wei Zhang, Dinggang Shen, Tianming Liu, and Shu Zhang.
\newblock A comprehensive review of multimodal large language models: Performance and challenges across different tasks, 2024.
\newblock URL \url{https://arxiv.org/abs/2408.01319}.

\bibitem[Wei et~al.(2022)Wei, Wang, Schuurmans, Bosma, Xia, Chi, Le, Zhou, et~al.]{wei2022chain}
Jason Wei, Xuezhi Wang, Dale Schuurmans, Maarten Bosma, Fei Xia, Ed~Chi, Quoc~V Le, Denny Zhou, et~al.
\newblock Chain-of-thought prompting elicits reasoning in large language models.
\newblock \emph{Advances in neural information processing systems}, 35:\penalty0 24824--24837, 2022.

\bibitem[xAI(2024)]{xai2024grok2}
xAI.
\newblock Grok-2 model card.
\newblock \url{https://x.ai/news/grok-2}, 2024.
\newblock Large language model.

\bibitem[Yin et~al.(2024)Yin, Fu, Zhao, Li, Sun, Xu, and Chen]{10.1093/nsr/nwae403}
Shukang Yin, Chaoyou Fu, Sirui Zhao, Ke~Li, Xing Sun, Tong Xu, and Enhong Chen.
\newblock A survey on multimodal large language models.
\newblock \emph{National Science Review}, 11\penalty0 (12):\penalty0 nwae403, 11 2024.
\newblock ISSN 2095-5138.
\newblock \doi{10.1093/nsr/nwae403}.
\newblock URL \url{https://doi.org/10.1093/nsr/nwae403}.

\bibitem[Yu et~al.(2016{\natexlab{a}})Yu, Poirson, Yang, Berg, and Berg]{yu2016modeling}
Licheng Yu, Patrick Poirson, Shan Yang, Alexander~C. Berg, and Tamara~L. Berg.
\newblock Modeling context in referring expressions.
\newblock \emph{arXiv preprint arXiv:1608.00272}, 2016{\natexlab{a}}.

\bibitem[Yu et~al.(2016{\natexlab{b}})Yu, Poirson, Yang, Berg, and Berg]{yu2016modelingcontextreferringexpressions}
Licheng Yu, Patrick Poirson, Shan Yang, Alexander~C. Berg, and Tamara~L. Berg.
\newblock Modeling context in referring expressions, 2016{\natexlab{b}}.
\newblock URL \url{https://arxiv.org/abs/1608.00272}.

\bibitem[Yuan et~al.(2024)Yuan, Duan, Blukis, Pumacay, Krishna, Murali, Mousavian, and Fox]{yuan2024robopoint}
Wentao Yuan, Jiafei Duan, Valts Blukis, Wilbert Pumacay, Ranjay Krishna, Adithyavairavan Murali, Arsalan Mousavian, and Dieter Fox.
\newblock Robopoint: A vision-language model for spatial affordance prediction for robotics.
\newblock \emph{arXiv preprint arXiv:2406.10721}, 2024.

\bibitem[Yuan et~al.(2025)Yuan, Duan, Blukis, Pumacay, Krishna, Murali, Mousavian, and Fox]{yuan2025robopoint}
Wentao Yuan, Jiafei Duan, Valts Blukis, Wilbert Pumacay, Ranjay Krishna, Adithyavairavan Murali, Arsalan Mousavian, and Dieter Fox.
\newblock Robopoint: A vision-language model for spatial affordance prediction in robotics.
\newblock In \emph{Proc. Conference on Robot Learning (CoRL)}, volume 270, pages 4005--4020, 2025.

\bibitem[Zhao et~al.(2024)Zhao, Zhang, Chia, Xu, Zhao, and Bing]{zhao2024autoarenaautomatingllmevaluations}
Ruochen Zhao, Wenxuan Zhang, Yew~Ken Chia, Weiwen Xu, Deli Zhao, and Lidong Bing.
\newblock Auto-arena: Automating llm evaluations with agent peer battles and committee discussions, 2024.
\newblock URL \url{https://arxiv.org/abs/2405.20267}.

\bibitem[Zheng et~al.(2023)Zheng, Chiang, Sheng, Zhuang, Wu, Zhuang, Lin, Li, Li, Xing, Zhang, Gonzalez, and Stoica]{zheng2023judgingllmasajudgemtbenchchatbot}
Lianmin Zheng, Wei-Lin Chiang, Ying Sheng, Siyuan Zhuang, Zhanghao Wu, Yonghao Zhuang, Zi~Lin, Zhuohan Li, Dacheng Li, Eric~P. Xing, Hao Zhang, Joseph~E. Gonzalez, and Ion Stoica.
\newblock Judging llm-as-a-judge with mt-bench and chatbot arena, 2023.
\newblock URL \url{https://arxiv.org/abs/2306.05685}.

\end{thebibliography}


\end{document}